# An entity-guided text summarization framework with relational heterogeneous graph neural network

Jingqiang Chen

School of Computer Science, Nanjing University of Posts and Telecommunications, Nanjing, China

cjq@njupt.edu.cn

ORCID: 0000-0001-7242-0141

**Abstract**

Two crucial issues for text summarization to generate faithful summaries are to make use of knowledge beyond text and to make use of cross-sentence relations in text. Intuitive ways for the two issues are Knowledge Graph (KG) and Graph Neural Network (GNN) respectively. Entities are semantic units in text and in KG. This paper focuses on both issues by leveraging entities mentioned in text to connect GNN and KG for summarization. Firstly, entities are leveraged to construct a sentence-entity graph with weighted multi-type edges to model sentence relations, and a relational heterogeneous GNN for summarization is proposed to calculate node encodings. Secondly, entities are leveraged to link the graph to KG to collect knowledge. Thirdly, entities guide a two-step summarization framework defining a multi-task selector to select salient sentences and entities, and using an entity-focused abstractor to compress the sentences. GNN is connected with KG by constructing sentence-entity graphs where entity-entity edges are built based on KG, initializing entity embeddings on KG, and training entity embeddings using entity-entity edges. The relational heterogeneous GNN utilizes both edge weights and edge types in GNN to calculate graphs with weighted multi-type edges. Experiments show the proposed method outperforms extractive baselines including the HGNN-based HGNNSum and abstractive baselines including the entity-driven SENECA on CNN/DM, and outperforms most baselines on NYT50. Experiments on sub-datasets show the density of sentence-entity edges greatly influences the performance of the proposed method. The greater the density, the better the performance. Ablations show effectiveness of the method.

## 1 Introduction

Automatic text summarization aims to distill long text into concise summaries to facilitate quick information consumption [1]. Extractive summarization directly extracts salient sentences from source documents as summaries. Abstractive summarization can generate text that does not appear in source documents. Recent progress of text summarization relies on the use of deep learning techniques [2-6]. These methods mainly follow the encoder-decoder framework where source texts are encoded by encoders in different forms, and sentences are labeled or generated by decoders.

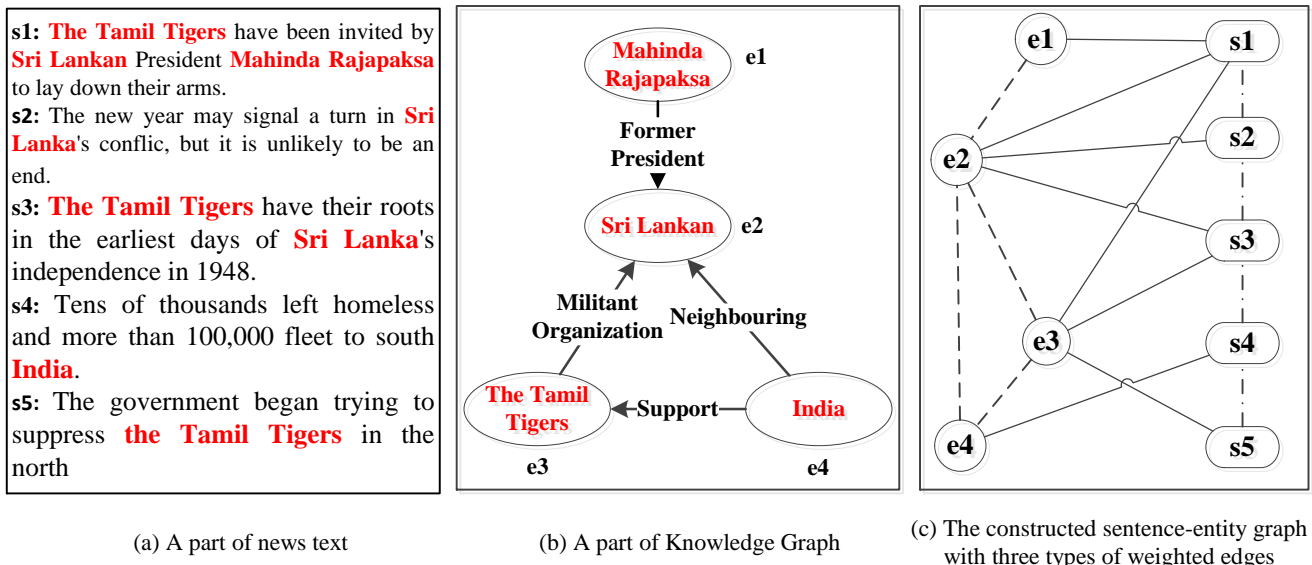

(a) A part of news text   (b) A part of Knowledge Graph   (c) The constructed sentence-entity graph with three types of weighted edges

F**ig. 1**  An example taken from CNN about the Tamil Tigers Organization of Sri Lankan

Two crucial issues for summarization to generate faithful summaries are: to make use of knowledge beyond text [1], and to make use of cross-sentence relations in text [2]. Fig. 1 shows an example. There are five sentences in Fig. 2 (a). The sentences s1 and s5 are relatively distant from each other, while there are cross-sentence relations between s1 and s2 because they contain the same entity Tamil Tigers. Fig. 1(b) is a part of Knowledge Graph about the entities recognized in the sentences. The knowledge graph is constructed outside the news text and contains knowledge beyond the text. Either cross-sentence relations shown in Fig. 1(a) or entity knowledge shown in Fig. 1(b) is important for summarization. A feasible way to utilize entities knowledge and cross-sentence relations is to construct a sentence-entity graph as in Fig. 1(c) and to link the graph to Knowledge Graph.

For the knowledge issue, knowledge for summarization is often beyond text [1, 7, 8], while most existing research is on text itself. Knowledge Graph is an important form of knowledge beyond text. Entities mentioned in text can be linked to entities in KG to make use of knowledge in text and knowledge beyond text. Entities in text contain knowledge such as keywords and structural knowledge in text, and can be used to improve summarization. For example, entities are used as keywords to generate entity-coherent summaries [3], and entities are used as bridge to cluster sentences and generate summaries for long Chinese articles [9]. To utilize entities in KG to improve summarization, a preliminary study simply injects entity embeddings trained on KG into the encoder-decoder architecture [10]. Recent work uses entities in KG to improve pre-trained language models [11, 12]. These previous studies utilize entities in text or entities in KG for summarization. It is still a challenge to effectively connect structural knowledge in text with knowledge in KG to improve summarization.

For the cross-sentence relations issue, modeling these relations is crucial to extracting summary-worth sentences which can be further summarized to abstractive summaries. Early traditional work such as LexRank [13] and TextRank [14] uses inter-sentence cosine similarity to build text graphs. Recent progresses are based on GNN [15] to capture long-distance dependency through modeling cross-sentence relations as graph structures. Various types of graph structures can be built, and different variations of GNNs are proposed to calculate the graphs. In particular for summarization, graphs constructed from text often contains real-value weighted multi-type edges. A recent study constructs a sentence-word bipartite graph with a single type of weighted edge, and a heterogeneous GNN is proposed to calculate node encodings [2]. GNN is also used to model intra- and inter-sentence relations for dialogue summarization [16]. Other work relies on discourse structures to build summarization graphs [17, 18]. For graphs with unweighted multi-type edges, the relational GNN (R-GNN) [19] is proposed to calculate the graphs. Edge types and edge weights are important information for GNN-based graph calculations, and Entities in text are informative semantic units for graph construction. Current studies for summarization do not make full use of entities, edge types, and edge weights for construction and calculation of graphs to model cross-sentence relations.

As an attempt for tackling the knowledge issue and the cross-sentence relation issue in combination for summarization, this paper proposes an entity-guided summarization framework by leveraging entities mentioned in text to connect KG and GNN, and by making use of edge weights and edge types in GNN for calculations of graphs with weighted multi-type edges.

Firstly, entities mentioned in text are used to build and calculate a sentence-entity graph with weighted multi-type edges to model sentence relations for summarization. Three edge types, i.e., sentence-entity edges, entity-entity edges, and sentence-entity edges are introduced to link sentences and entities. These edges have different weights. The relational heterogeneous GNN (R-HGNN) for summarization is proposed for calculations of the graph by making use of edge types and edge weights in the propagation process. R-HGNN combines the advantage of the traditional GNN [15] in making use of edge weights and the advantage of the traditional R-GNN [19] in making use of edge types.

Secondly, entities are used to link the sentence-entity graph to Knowledge Graph to utilize knowledge outside text for summarization using entity linking techniques [20-22]. It is reasonable to believe that external knowledge can improve summarization. Knowledge Graph is utilized in GNN in three ways: building entity-entity edges as cooccurrence times of two entities in Wikipedia webpages, initializing entities embeddings in knowledge bases by RDF2Vec [23], and supervising entities embeddings with entity-entity edges.

Thirdly, entities guide the two-step summarization method. A multi-task objective is defined on the sentence-entity graph network to select salient sentences and entities simultaneously. Sentence selection and entity selection can benefit each other, because salient sentences often contain salient entities. An entity-focused abstractor follows to compress the salient sentences using the salient entities as queries.

The main contributions of this paper are summarized as follows:

- To the authors' knowledge, this paper is the first to leverage entities to connect GNN and Knowledge Graph to model cross-sentence relations in text and knowledge beyond text for summarization. GNN is connected with KG by constructing the sentence-entity graph where entity-entity edges are built based on KG, initializing entity embeddings on KG, and training entity embeddings using entity-entity edges.
- The relational heterogeneous GNN for summarization is proposed to calculate graphs with weighted multi-type edges by making use of both edge weights and edge types in the propagation process.
- The proposed method outperforms all existing baselines on the CNN/DM dataset without the pre-trained language models, and outperforms most baselines on the NTY50 dataset. Experiments on sub-datasets show the density of sentence-entity edges greatly influence the performance of the proposed method. Greater the density of sentence-entity edges, better the performance of the method. Ablation studies show the effectiveness of the proposed method.

## 2 Related work

**Entities and knowledge graphs for summarization** Entity plays an important role in summarization for selecting sentences [3, 24, 25]. Entities connecting sentences are used to extract coherent sentences [26]. More recently, entities are used to select summary-worth sentences by computing an entity context vector which is compared with the sentence context vectors, and are utilized to generate coherent abstractive summaries by compressing the extracted sentences [3]. And named entities extracted from input documents are used to construct entity-predicate-entity graphs, and then the graph2seq method is employed to generate summaries [27]. Knowledge Graph is an important form of knowledge beyond text for summarization. Another similar work is Semantic Link Network which can be traced to the work of constructing a network models in 1998 and is recently extended to cyber-physical-social space for better modeling cyber-physical-social systems [28]. Most previous studies make use of KG by training entity embeddings based on KG. For example, entity-level knowledge from knowledge graphs are incorporated into the encoder-decoder architecture for summarization [10] by learning entity embeddings through the TransE method [29] and then injecting the entity embeddings into Transformer [30]. With the development of the pre-trained language models, entity-level knowledge can also be used to improve BERT [31] such as ERNIE [11]. The K-Bert model is proposed to inject domain knowledge into BERT through connecting triplets with entities in a sentence to construct a sentence tree [12]. These entity embeddings trained based on KG reflect the structure of entities and relations in KG. For summarization, the cross-sentence structure in the document is also important. It is a promising research front for summarization to make use of knowledge of entity relations in KG and knowledge of cross-sentence relations in text in combination. This paper proposes to inject knowledge in KG into the sentence-entity graph by building entity-entity edges based on KG, initializing entity embeddings on KG, and training entity embeddings using entity-entity edges.

**GNNs for summarization** Graph neural networks are originally designed for homogeneous graph with nodes of same types and with weighted edges [32, 33]. Message propagation algorithms for the traditional GNNs utilize the weighted edges to propagate message from neighboring nodes to calculate node encodings [33]. The original GNN does not consider edge types. Therefore, the relational GNN (R-GNN) is proposed to utilize the edge types by equipping each edge type with a transformation function [19]. The original R-GNN does not consider edge weights. GNNs are effective approach to model the structure of documents and to capture long-distance relationships in text for summarization. For example, the GNN is employed for extractive summarization by constructing the sentence graph with sentences encoded by RNN and edge weights computed by counting discourse relation indicators, and then applying the traditional GNN to calculate the graph [17]. The R-GNN-based summarization approach is is proposed to capture long-distance relationships in long text by constructing the graph on sentences, words and entities with the relations of NEXT, IN and REF [34]. The GNN is used in a discourse-aware neural summarization model to capture long-range dependencies by constructing structural discourse graph based on RST trees and coreference mentions encoded with GNN [18]. Recently, the heterogeneous GNNs with different types of nodes and edges are proposed for summarization [2] or other applications [35-40]. For summarization, the heterogeneous graph neural network (HGNN) is proposed by constructing a sentence-word bipartite graph having only one type of edge to model cross-sentence relations [2], and applying GAT [32] for node encodings calculations. For other applications, the heterogeneous GNNs are applied to recommendation [34], or to program reidentification [37]. For large-scale graphs, HinSAGE [41] is an extension of GraphSAGE [38] to heterogeneous networks, with three types of aggregating algorithms i.e. mean aggregating, LSTM aggregating, pooling aggregating. Most existing heterogeneous GNNs are for graphs with 0/1 edges (un-weighted edges). To calculate graphs with weighted multi-type edges, both edge weights and edge types should be made use of in the propagation process. This paper proposes the relational heterogeneous GNN for summarization to make use of both edge weights and edge types by combining the advantages of both the traditional GNN and the traditional R-GNN.

**Two-step summarization approaches** With the development of deep learning techniques, great progress has been made in extractive and abstractive summarization. Most work focuses on the encoder-decoder model based on RNN [4, 5, 42] or Transformer [6, 43]. To connect the extractive and abstractive summarization, the two-step summarization approach has been popular in recent years [3, 44]. The extraction step selects summary-worth sentences, and the abstraction step generates the abstractive summary from the selected sentences [45-47]. Some work applies the reinforcement method to connect the two steps and to enhance the ROUGE gains [3, 9, 48]. In [3], entities are leveraged to select sentences for abstraction. In [9], GNN is used to select sentence cluster for abstraction by first clustering sentences containing same keywords and then linking the clusters. This paper proposes the multi-task selector to select salient sentences and salient entities, and employs the entity-focused abstractor to compress the sentences.

## 3 The proposed framework

The goal is to create both extractive and abstractive summaries for an input document by making use of knowledge beyond text and cross-sentence relations. The input document is denoted as *D*, and has *M* sentences and *N* entities. Each entity may be mentioned many times in the document with different mention forms. For example, the entity Tamil Tigers in the example in Fig. 1 may be mentioned as Tigers, Liberation Tigers of Tamil Eelam, LTTE, The Militant Organization, etc. Therefore, an entity in a document is represented as {EntityName, MentionSet}, where EntityName is the entity name, and MentionSet is the set of mentions of the entity in the document. Each sentence is a sequence of words, and each entity mention is also a sequence of words.

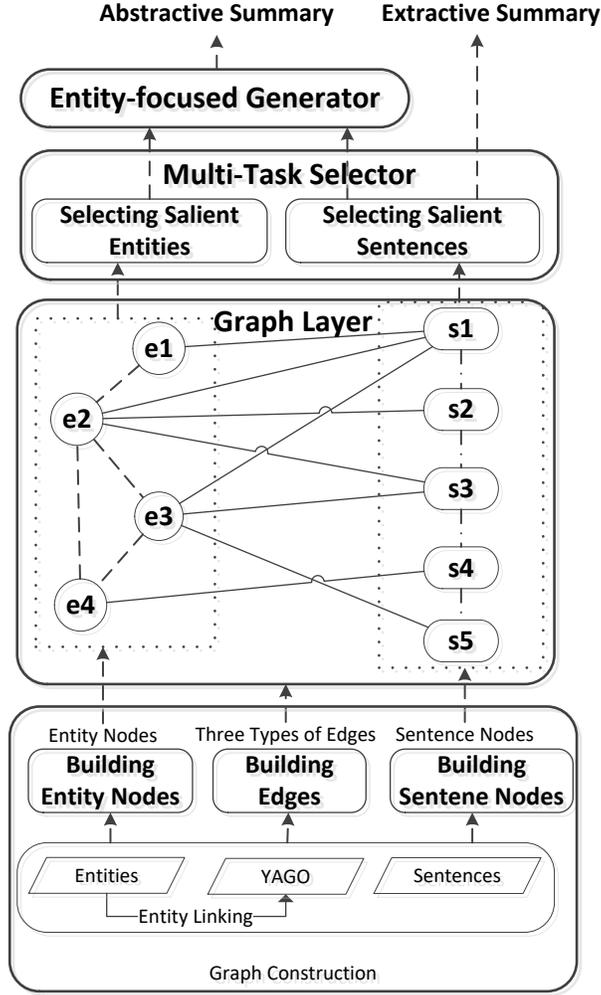

F**ig. 2** The framework of the proposed entity-guided summarization model

To create both extractive and abstractive summaries, the two-step summarization approach is employed to select salient sentences and salient entities followed by a generator to create abstractive summaries. And a multi-task objective is defined for salient sentence selecting and salient entity selecting, because salient sentences often contain salient entities. Sentence selection is formulated as a sequence labeling task [5], as well as entity selection. A label sequence $y^S_1, y^S_2, \ldots, y^S_M$ is predicted where $y^S_i=1$ denotes the $i^{th}$ sentence is selected as the summary sentence and $y^S_i=0$ denotes the sentence is not selected. A label sequence $y^E_1, y^E_2, \ldots, y^E_N$ is predicted where $y^E_j=1$ denotes the $j^{th}$ entity is selected as the salient entity and $y^E_j=0$ denotes the entity is not selected. The ground truth sentence labels (called ORACLE) are obtained using the greedy approach introduced in [3], and the ground truth entity labels are obtained by collecting entities in the ground truth manual abstractive summaries. The abstractive summary generator is a seq2seq network equipped with the attention mechanism [48] and the copy mechanism [49], using selected entities as queries.

To make use of external knowledge and cross-sentence relations, entities serve as guidance in the proposed framework. Entities guide the construction and calculation of the sentence-entity graph where sentences are linked with each other through entities, and entities mentioned in text are linked to YAGO2 [50] to collect external knowledge for summarization. The relational heterogeneous graph neural network (R-HGNN) is proposed to calculate node encodings of the sentence-entity graph for summarization.

Fig. 2 shows the proposed framework. As a concrete example, the sentence-entity graph in Fig. 2 is constructed from the example in Fig. 1. The graph comprises entity nodes and sentence nodes, connected by three types of relational edges which are differently shaped in Fig. 2. The proposed framework consists of: 1) the graph constructing module builds and initializes the sentence-entity graph for an input document (§3.1), 2) the graph layer applies the relational heterogeneous GNN to calculate node encodings (§3.2), 3) the multi-task selector that select salient sentences and entities by training R-HGNN using a multi-task objective (§3.3), and 4) the generator that generates abstractive summaries by the entity-focused pointer-generator network taking the selected sentences and entities as input (§3.4). Finally, an RL connector connects the selector and the generator (§3.5). Due to the limited computational resource, pre-trained encoders (i.e., BERT) are not applied, which are regarded as future work. Moreover, the proposed model is orthogonal to BERT-based models.

**3.1 Constructing a sentence-entity graph for a document**

Given a document, a heterogeneous sentence-entity graph is constructed. Previous work [2] uses sentences and words as nodes to build heterogeneous bipartite graph which has only a single type of edges i.e. the sentence-word edge. This research uses entities instead of words as the nodes, because: 1) entities are more informative than words, and more importantly, 2) and entities mentioned in text can be linked to Knowledge Graphs to make use of knowledge beyond text. However, as every coin has two sides, entities are sparser than words in text. Therefore, apart from sentence-entity edges, sentence-sentence edges and entity-entity edges are added to the sentence-entity graph as shown in Fig. 2. Each sentence corresponds to a sentence node in the graph, and each entity in the document corresponds to an entity node in the graph. Nodes and edges are built and encoded in the following. Knowledge Graph plays an important part in entity encoding and entity-entity edges constructing.

**Building entity nodes** Since each entity may have many mentions in the document, the off-the-shelf NER tool is employed to recognize entity mentions in the input document, and then the off-the-shelf coreference resolution tool is used to cluster the mentions. Both tools are from Stanford CoreNLP [51]. Technically, the entity mentions recognized by the NER tool are precise enough, but the number of recognized mentions is limited. Therefore, the coreference resolution tool is used to recall more mentions and to cluster the mentions based on the reference chains. To balance the precision and the mention number, only the chains that containing mentions recognized by the NER tool are used. A mention cluster corresponds to an entity. Then, entities nodes are built and encoded as shown in Fig. 3.

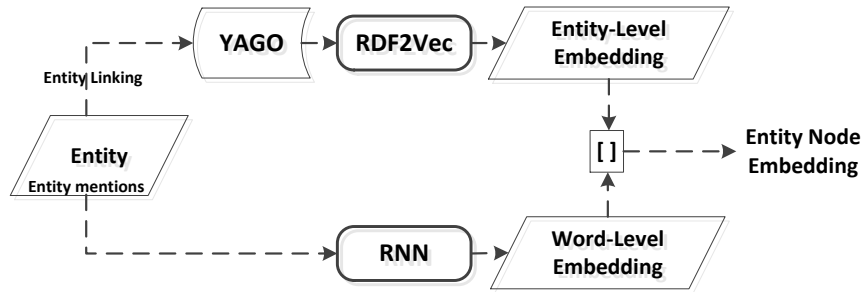

F**ig. 3** Building and encoding entity nodes

In Fig. 3, to connect the sentence-entity graph with Knowledge Graph, entities in the input document are linked to YAGO2 with entity linking techniques. YAGO2 is an open source knowledge base extracted from Wikipedia, comprising a huge number of entities and facts [50]. Entity linking aims to link entities mentioned in text to entities registered in the knowledge base. Entity linking is a hard job because entities have ambiguous mentions in text. Recent years witness numerous studies on entity linking with the development of Knowledge Graph and deep learning techniques [20-22]. Among them, this study adopts the state-of-the-art open-source framework AIDA-light [21] to link each entity mention cluster to entities registered in YAGO2. AIDA-light is a lightweight collective entity linking framework, unifying three features, i.e. the prior probability of an entity being mentioned, the similarity between the contexts of a mention and a candidate entity, as well as the coherence among entities.

In Fig. 3, entity node embeddings are obtained by concatenating word-level embeddings and entity-level embeddings. 1) Word-level entity embeddings reflect word-level literal meanings of entities. The mentions in a mention cluster are ordered as they occur in the input document and are then concatenated into one sequence with the special token <SEP> as the separator. A Bi-directional Recurrent Neural Network (BiRNN) is employed to encode each mention sequence. The last forward hidden state and the last backward hidden state are concatenated as the word-level entity embedding of the $j^{th}$ entity denoted as $e_j^W$. 2) Entity level embeddings reflect entity relatedness in Knowledge Graph. RDF2Vec [23] is adopted to learn entity-level embeddings from YAGO2. RDF2Vec first converts the knowledge graph into a set of sequences of entities using graph walks, and then uses those sequences to train a neural language model estimating the likelihood of a sequence of entities appearing in a graph. The learned entity-level entity embedding of the $j$th entity is denoted as $e_j^E$, and entity-level embeddings of all entities in the document are denoted as the matrix $E^E$ of which the $j$th row equals with $e_j^E$. Since the total number of entities in YAGO2 is rather huge and it is also unnecessary to learn embeddings for all entities in YAGO2, this study maintains entity-level entity embeddings for a fixed-sized entity vocabulary of which each entity occurs frequently in the experimental corpora. The special entity UNK is used for entities that are out of the vocabulary. Word-level embeddings $e_j^W$ and entity-level embedding $e_j^E$ are concatenated to serve as initial encodings of the $j$th entity node in the sentence-entity graph, denoted as $E_j^{(0)} = [e_j^W, e_j^E]$. Take the entity Tamil Tigers as example. It is mentioned in the document as Tigers, Liberation Tigers of Tamil Eelam, LTTE, The Militant Organization. These mentions are concatenated with <SEP> and are encoded with BiRNN to get word-level embeddings. This entity is also in YAGO2 and can be encoded with RDF2Vec to get entity-level embeddings. Word- and entity-level embeddings are concatenated as the final embeddings of the entity. This way entity nodes can embrace word-level lexical information and entity-level information from Knowledge Graph.

**Building sentence nodes** The $M$ sentences of the input document is represented as $[s_1, s_2, s_3, …, s_M]$. The sentence $s_i$ is a sequence of words represented as $[s_{i1}, s_{i2}, s_{i3}, …, s_{i|si|}]$, where $|si|$ is the length of the $i$th sentence $s_i$. Then a two-level BiRNN is applied to encode the sentence sequence as shown in Fig. 4. Equation (1) employs the word-level BiRNN to encode each sentence word by word, and equation (2) concatenates the last forward hidden state and the last backward hidden state as the sentence representation. Equation (3) employs the sentence-level BiRNN taking sentence representations as input to encode the sentence sequence to conserve sequential relationships of sentences, and then equation (4) concatenates the forward hidden state and the backward hidden state as the initial sentence node representation $S_i^{(0)}$. As an example, the five sentences in Fig. 1 (a) are first encoded with equation (1) and (2) independently, and are then encoded with equation (3) and (4) sequentially to get final sentence representations.

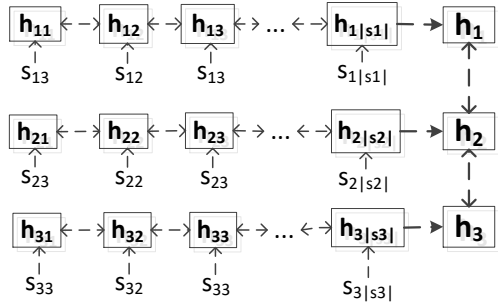

**Fig. 4** Encoding sentences

$$[\vec{h}_{i1}, \vec{h}_{i2}, ..., \vec{h}_{i|si|}; \overleftarrow{h}_{i1}, \overleftarrow{h}_{i2}, ..., \overleftarrow{h}_{i|si|}] = BiRNN1([s_{i1}, s_{i2}, ..., s_{i|si|}]) \quad (1)$$

$$s\_rep_i = [\vec{h}_{i|si|}, \overleftarrow{h}_{i1}] \quad (2)$$

$$[\vec{h}_1, \vec{h}_2, ..., \vec{h}_M; \overleftarrow{h}_1, \overleftarrow{h}_2, ..., \overleftarrow{h}_M] = BiRNN2([s\_rep_1, s\_rep_2, ..., s\_rep_M]) \quad (3)$$

$$S_i^{(0)} = [\vec{h}_i, \overleftarrow{h}_i] \quad (4)$$

**Building edges** There are three types of edges in the sentence-entity graph. 1) The sentence-entity (SE for short) edge is built between a sentence node and an entity node if the sentence contains any mention of the entity. The SE edges reflect containment relationships between sentences and entities, and the weights are set as the occurrence times of the entity in the sentence. 2) The sentence-sentence (SS for short) edge is built between every two adjacent sentences in the input document, so there are $M-1$ SS edges in the graph of the document of M sentences. The SS edges reflect sequential relationships among sentences, and the weights are set as 1. 3) The entity-entity (EE for short) edge is built between any two entities if the two entities co-occur in a Wikipedia webpage. The EE edge is built based on the external knowledge base, different from other two types of edges that are built based on the input document itself. Since entities registered in YAGO2 are from Wikipedia, each entity corresponds to a Wikipedia webpage. The weights of EE edges are set as the numbers of Wikipedia webpages that two entities co-occur. Fortunately, the data of entity-entity cooccurrences in Wikipedia are provided in AIDA. As the example in Fig. 2 shows, there is an EE edge between the entities e2 and e3 because the two entities occur in a same Wikipedia webpage. There is an SE edge between the sentence s3 and the entity e2 because s3 contain e2 as shown in Fig.1 (a). And there is an EE edge between the sentences s2 and s3 because s2 and s3 are adjacent sentences. The three edge types represent three orthogonal relationships, and are all undirected edges.

Entity-entity edges are built based on KG and Entity-level entity embeddings are initialized on KG. In the following sections, the training of the entity encodings will be supervised using entity-entity edges.

## 3.2 The relational heterogeneous GNN for summarization

This section introduces the relational heterogeneous graph neural network to compute node encodings for the sentence-entity graph with weighted multi-type edges for summarization. Edge types and edge weights are both important information for message propagation in GNN. The traditional GNN [15] makes use of edge weights to calculate graphs with weighted single-type edges, and the traditional R-GNN [19] makes use of edge types to calculate graphs with unweighted multi-type edges. For the sentence-entity graph with weighted multi-type edges, R-HGNN makes use of both edge types and edge weights by combining the advantages of both GNNs. R-HGNN employs the propagation algorithm of GNN for intra-edge-type propagation to calculate node encodings, defines an edge-type-specific function as R-GNN does to transform the encodings to edge-type-specific encodings, and aggregates encodings.

**Definition of R-HGNN** Suppose the input graph has m nodes and c edge types. The edge types are denoted as $\{ET_1, ET_2, \ldots, ET_c\}$. Nodes are linked by different types of weighted edges, and each edge type corresponds to an independent adjacent matrix of Nodes. The notations are as follows:

- $X^{(l)} \in R^{m \times d}$ is the node encodings in the $l^{th}$ level of R-HGNN, and $X^{(0)}$ is the initial node embeddings.
- $A^{ET_k} \in R^{m \times m}$ is the edge-type-specific adjacent matrix of nodes for the edge type $ET_k$, and elements in $A^{ET_k}$ are the weights of corresponding edges.

The goal of R-HGNN is to learn a function $Z = f(X^{(0)}, A^{ET_1}, A^{ET_2}, \ldots, A^{ET_c})$ where Z is the high-level hidden features for the nodes, encapsulating the information of edge types, edge weights, and the graph structure. R-HGNN has $L$ levels, and $Z = X^{(L)}$. Equations (5) to (7) are the equations for the propagation process of R-HGNN to calculate node encodings in the $l^{th}$ level.

$$X^{ET_k^{(l)}} = D^{ET_k^{-\frac{1}{2}}} A^{ET_k} D^{ET_k^{-\frac{1}{2}}} g^{ET_k}(X^{(l-1)}) \quad (5)$$

$$X^{self^{(l)}} = g^{self}(X^{(l-1)}) \quad (6)$$

$$X^{(l)} = \sigma(\sum_{k=1}^{c} X^{ET_k^{(l)}} + X^{self^{(l)}}) \quad (7)$$

Equation (5) calculates edge-type-specific nodes encodings by making use of both edge weights and edge types. Firstly, edge weights are used in the propagation process by partly borrowing the idea of the original GNN proposed in [15]. As with [15], the edge-type-specific matrix $A^{ET_k}$ is normalized by the degree matrix $D^{ET_k}$ where the diagonal element $D_{ii}^{ET_k} = \sum_j A_{ij}^{ET_k}$, because directly using the edge-type-specific matrices will change the scale of node encodings. The theoretical justification of the calculations is provided in [15]. Secondly, edge types are used in the propagation process by partly borrowing the idea of the original R-GNN proposed in [19]. As with [19], each edge type is equipped with an edge-type-specific transformation function $g^{ET_k}()$, and the linear transformation function $g^{ET_k}(X) = W^{ET_k} X$ with a trainable weight matrix $W^{ET_k}$ is chosen. The node encodings are transformed by the function $g^{ET_k}()$, and are propagated to neighboring nodes through the edge-type-specific propagation process. The virtual self-edges are added to the graph to consider self-loops of the propagation process, and equation (6) calculates node encodings for the self-edges.

Equation (7) calculates encodings of each node in the $l^{th}$ level by accumulating all edge-type-specific encodings of the node, where $\sigma()$ is the element-wise activation function such as $\operatorname{Re}LU(\cdot) = \max(0, \cdot)$. The accumulated encodings in the $l^{th}$ level are used for propagation in the $(l+1)^{th}$ level. And the encodings in the final $L^{th}$ level is used as features of the nodes in the next subsection.

R-HGNN utilizes the advantages of both the original GNN and the original R-GNN to make use of edge types and edge weights for graphs with weighted multi-type edges. To some degree, the original GNN and the original R-GNN can be seen as the special case of R-HGNN without considering edge types and without considering edge weights respectively.

**Applying R-HGNN to the sentence-entity graph** The sentence-entity graph constructed in the previous subsection has three types of weighted edges, i.e., the SE edges, the SS edges, and the EE edges. The graph has $M$ sentence nodes and $N$ entity nodes. The proposed R-HGNN can be applied to calculate node encodings. The inputs are formalized as follows:

- $A^{SS} \in R^{(M+N) \times (M+N)}$, the adjacent matrix for the SS edges, where $A_{ij}^{SS} = 1$ if $i$ and $j$ are both sentence nodes and are adjacent in the document, else $A_{ij}^{SS} = 0$, as described in Section 3.1.
- $A^{SE} \in R^{(M+N) \times (M+N)}$, the adjacent matrix for the SE edges, where the value of $A_{ij}^{SE}$ is set as occurrence times of entities in sentences if $i$ and $j$ are different types of nodes, else $A_{ij}^{SE}$ is set as 0, as described in Section 3.1.
- $A^{EE} \in R^{(M+N) \times (M+N)}$, the adjacent matrix for the EE edges, where the value of $A_{ij}^{EE}$ is set as the co-occurrence times of the entity $i$ and the entity $j$ in Wikipedia Web Pages if $i$ and $j$ are both entity nodes, else $A_{ij}^{EE}$ is set as 0, as described in Section 3.1.

- $X^{(0)} = \begin{bmatrix} S^{(0)} \\ E^{(0)} \end{bmatrix} \in R^{(M+N) \times d}$, the initial node embedding matrix, where $S^{(0)} \in R^{M \times d}$ is the matrix of initial sentence node encodings and $E^{(0)} \in R^{N \times d}$ is the matrix of initial entity node encodings, calculated as described in Section 3.1. $X^{(0)}$ is the concatenation of $S^{(0)}$ and $E^{(0)}$.

The outputs are the node encodings in the $L^{th}$ level of R-HGNN, i.e., the sentence node encodings $S^{(L)}$ and the entity node encodings $E^{(L)}$. Take the graph in Fig. 2 as an example. There are five sentences and four entities in the graph. The adjacent $A^{SS}$, $A^{SE}$ and $A^{EE}$ can be computed accordingly, and the node embedding matrix $X^{(0)}$ can also be initialized as described in Section 3.1. R-HGNN can be applied in the graph to calculate sentence node encodings and entity node embeddings. In the following subsection, a multi-task selector is defined upon the R-HGNN, using these encodings as features.

### 3.3 The multi-task selector

The multi-task selector has two tasks to do: selecting salient sentence nodes and select salient entity nodes from the sentence-entity graph. The two tasks can benefit each other, because salient sentences often contain salient entities. As the example in Fig. 1 (a) shows, the leading sentence is a salient sentence, and contains the salient entity Tamil Tigers. Selection of salient sentences and entities can affect each other. The selected sentences form the extractive summary, and the selected entities can used as queries of the entity-focused generator introduced in next subsection. With node encodings of the last level of the R-HGNN as features, the objectives of the multi-task selector are defined as follows:

**Supervising with sentence selecting and entity selecting collectively** Equations (8) computes the probability of the sentences to be selected by applying a softmax function over sentence node encodings with MLP transformations. Equation (9) computes the probability of the entities to be selected by applying a softmax function over entity node encodings with MLP transformations.

$$p(\hat{y}_i^S = 1) \sim soft\max(MLP(S^{(L)})) \quad (8)$$

$$p(\hat{y}_j^E = 1) \sim soft\max(MLP(E^{(L)})) \quad (9)$$

Cross Entropy is adopted to compute the loss. As mentioned, the label $y_i^S = 1$ means that the $i^{th}$ sentence is a summary sentence and $y_i^S = 0$ means the sentence is not a summary sentence, so the ground-truth probability that the $i^{\text{th}}$ sentence to be selected is $p(y_i^S = 1) = \frac{y_i^S}{\sum_{i=1}^{M} y_i^S}$. The entity label $y_j^E = 1$ means that the $j^{\text{th}}$ entity is a summary entity and $y_i^E = 0$ means the entity is not a summary entity, so the ground-truth probability that the $j^{\text{th}}$ entity to be selected is $p(y_i^E = 1) = \frac{y_j^E}{\sum_{j=1}^{N} y_j^E}$. Equations (10) and (11) compute the sentence loss and the entity loss respectively.

$$loss^S = CrossEntropy(y^S, \hat{y}^S) \quad (10)$$

$$loss^E = CrossEntropy(y^E, \hat{y}^E) \quad (11)$$

**Supervising entity embeddings with entity-entity edges** Entity embeddings consist of word-level embeddings and entity-level embeddings, where entity-level embeddings are initialized by RDF2Vec to reflect the global relatedness of entities in YAGO2. Since the embeddings are trainable in the proposed model, the relatedness information between entities will be lost or reduced in entity embeddings if the training of entity embeddings is not directly supervised. Moreover, the local entity relatedness reflecting the structure of the constructed sentence-entity graph should also be enhanced in the entity embeddings. It is reasonable to believe that keeping entity relatedness information in entity embeddings can improve the performance of the sentence selector and the entity selector.

$$r_{i,j}^{EE} = \frac{A_{i,j}^{EE}}{\sum_{i,j} A_{i,j}^{EE}} \quad (12)$$

$$\hat{r}_{i,j}^{EE} \sim soft\max(E^E \times E^{E^T}) \quad (13)$$

$$loss^{EE} = CrossEntropy(r^{EE}, \hat{r}^{EE}) \quad (14)$$

Because entity-entity edges reflect entity relatedness information in Wikipedia, entity-entity edges are used as ground-truth entity-entity relatedness to supervise the training of entity embeddings. Firstly, entity-entity edges for every two entities in the document are normalized by equation (12), where $A^{EE}$ is the adjacent matrix for EE edges, and the element $A_{i,j}^{EE}$ of the matrix is the co-occurrence times of the entity i and the entity j in Wikipedia webpages as calculated in section 3.1. Only entities in the document are considered, because 1) the entities co-occur in the same document have higher relatedness, which will be reflected in the entity embeddings; 2) using a small entity set for supervising instead of the large entire entity set can save the limited computational resource and speed up training. Secondly, entity-entity relatedness is predicted by using the softmax activation function to normalize the dot-product between the entity-level entity matrix $E^E$ and $E^{E^T}$ in equation (13), where $E^E$ is the matrix of the entity-level entity embeddings of entities in the input document as calculated in section 3.1. Finally, Equation (14) computes the loss as the cross entropy between the ground-truth and the predicted entity-entity relatedness.

**Loss of the multi-task selector** The final loss of the multi-task selector is defined in equation (15) by linearly combining the sentence selecting loss, the entity selecting loss, and the entity-entity relatedness loss. The hyper-parameters $\lambda^E$ and $\lambda^{EE}$ are empirically set as 0.42 and 0.33 respectively.

$$loss^{selector} = loss^S + \lambda^E * loss^E + \lambda^{EE} loss^{EE} \quad (15)$$

**Inference** In the inference stage, sentences and entities are ranked by $P(\hat{y}_i^S = 1)$ and $P(\hat{y}_i^S = 1)$ respectively, and then the top-ranked ones are selected.

### 3.4 The entity-focused generator

The entity-focused generator inputs the salient sentences and salient entities selected by the multi-task selector, and generates abstractive summaries. The entity-focused generator extends the state-of-the-art pointer-generator network [49] through using the entities as queries.

The sentences are ordered as they occur in the original document, and are concatenated to a whole text. Then a BiRNN is applied to encode the text. Let $\vec{h}_i^T$ and $\overleftarrow{h}_i^T$ be the forward and backward hidden states of the $i^{th}$ word in the concatenated text. $d\_rep = [\vec{h}_m^T, \overleftarrow{h}_1^T]$ is the representation of the concatenated text where m is the word count of the text. Let $h_i^T = [\vec{h}_i^T, \overleftarrow{h}_i^T]$ be the encoding of the $i^{th}$ word.

For entity encodings, only the word-level entity embeddings (denoted as $e_j^W$ for the $j^{th}$ entity as described in Section 3.1) are adopted to represent the entity. Then the average pooling is applied over the encodings of the entities to get the encoding of the salient entity set as $h^E = avg_{j=1}^N(e_j^W)$.

As with the original pointer-generator network, the decoder generates words following the vocabulary distributions and pointer distributions of the words as shown below.

$$p(w_t) = p_{gen} p_{vocab}(w_t) + (1 - p_{gen}) \sum_{i:w_i=w_t} a_{t,i} \quad (16)$$

The salient entities are used to calculated the attention $a_{t,i}$ and the generation probability $p_{gen}$. Let $h_t$ be the hidden state in the $t^{th}$ decoding step. $p_{gen}$ is calculated as follows:

$$a_t = soft\max(\alpha_t) \quad (17)$$

$$\alpha_{t,i} = v\tanh(W^{aD}h_t + W^{aT}h_i^T + W^{aE}h^E + b^{attn}) \quad (18)$$

$$p_{gen} = sigmoid(W^{pD}h_t + W^{pT}h_i^T + W^{pE}h^E + x_t + b^{attn}) \quad (19)$$

In the equations, the entity encodings are added as parameters to compute the attention and the generation probability. This way entity information is incorporated to generate summaries. As with the pointer-generator network, the coverage loss $cov\_loss_t$ is also added.

The loss is the negative log-likelihood of the predicted word and the coverage loss, i.e., $loss_t^{generator} = -\log(p(w_t)) + \lambda^{cov} cov\_loss_t$. Readers can refer to [49] for more details.

### 3.5 Connecting the selector and the generator

The selector selects salient sentences and salient entities, whereas the generator compresses and paraphrases them. Until this point, they are trained separately without any form of parameter sharing. To connect two networks, the self-critical learning algorithm [52] based on policy gradient is adopted to connect them.

In line with the Markov Decision Process formulation, at each time step, the selector samples sentences and entities from the input document, and the generator uses the sentences and entities to generate an abstractive summary. This summary is evaluated against the ground-truth summary, and receives ROUGE-1 [53] as the reward.

Following [52], in the training process, the selector samples sentences and entities from the distributions $p(\hat{y}_i^S = 1)$ and $p(\hat{y}_j^E = 1)$ introduced in Section 3.3. The generator then creates an abstractive summary and return the Rouge-1 reward denoted as $R$. Let $Sample^S$ be the index set of the sampled sentences in the input document, and $Sample^E$ be the index set of the sampled entities. For each $i \in Sample^S$, $y_i'^S$ is set as 1 and $p(y_i'^S = 1) = \dfrac{y_i'^S}{\sum_{k \in Sample^S} y_k'^S}$, and for $i \notin Sample^S$, $p(y_i'^S = 1) = 0$. For each $j \in Sample^E$, $y_j'^E$ is set as 1 and $p(y_j'^E = 1) = \dfrac{y_j'^E}{\sum_{k \in Sample^E} y_k'^E}$, and for $j \notin Sample^E$, $p(y_j'^E = 1) = 0$.

As with previous work [3], the parameters of the generator are frozen. Only the selector is trained by reinforcement learning. The following equations compute the loss of reinforcement learning.

$$loss^{RL} = R*(CrossEntropy(p(y'^S), p(\hat{y}^S)) + \lambda^E CrossEntropy(p(y'^E), p(\hat{y}^E))) \quad (20)$$

If the selector accurately selects salient sentences and entities, the entity-focused generator is more likely to generate high-quality abstractive summaries, which will be encouraged. Otherwise, actions resulting in inferior selections will be discouraged.

Equation (21) recomputes the loss of the selector with the RL loss, where the hyper-parameter $\lambda^{RL}$ is empirically set as 0.6.

$$loss^{selector\_with\_RL} = loss^{selector} + \lambda^{RL} loss^{RL} \quad (21)$$

## 4 Experiments

### 4.1 Corpora and preprocessing

Two popular summarization datasets are used to evaluate the proposed model: The CNN/DailyMail (CNN/DM) dataset [54] and the NYT50 dataset [55].

For CNN/DM, the standard dataset is split into 286649/13359/11490 examples for training, validation, and test. The preprocess steps in [54] are followed to obtain the plain text. To obtain entities and mentions, the annotations provided by the original dataset are used. For each entity in an example, the dataset provides the entity id and all mention starts and ends.

NYT50 is a subset of New York Times Annotated Corpus [56] preprocessed by [55] for document summarization. The dataset contains 110540 articles with summaries and is split into 100834 and 9706 for training and test. Following the preprocessing steps in [55], the last 4000 examples of the training set are used as the validation set, and the test examples are filtered to 3452. To obtain entities and mentions, the standoff annotations that contain entity mention annotations and coreference annotations provided in NYT50 are used. Mentions are clustered by coreference ids in the annotation file, and each mention cluster represents an entity. Entity types are also provided in annotation files. Entities of the location type, the person type, the organization type and the event type are used for experiments.

**Table 1** The statistics of the two datasets, where Doc.Sent, Sum.Sent, Doc.Ent, Sum.Ent, Sent.Men, Ent.YAGO, and SE.Density stand for the average sentence number in documents, the average sentence number in summaries, the average entity number in documents, the average entity number in summaries, the average entity mention number in sentences, the average number of entities successfully linked to YAGO2, and the density of sentence-entity edges respectively

|  | Train | Dev | Test | Doc.Sent | Sum.Sent | Doc.Ent | Sum.Ent | Sent.Men | Ent.YAGO | SE.Density |
|---|---|---|---|---|---|---|---|---|---|---|
| **CNNDM** | 286649 | 13359 | 11490 | 27.93 | 3.75 | 22.59 | 3.64 | 2.14 | 16.81 | 0.94 |
| **NYT50** | 100834 | 4000 | 3452 | 40.81 | 2.94 | 21.21 | 2.59 | 0.88 | 8.75 | 0.57 |

Table 1 shows statistics of the two datasets. The observations are as follows:

- Though CNN/DM have less sentences in documents than NYT50, CNN/DM has more entities than NYT50, which indicates CNN/DM is entity-denser than NYT50.
- Sentences in CNN/DM contain average 2.14 entity mentions, while sentences in NYT50 contain only average 0.88 mentions, indicating that sentences in CNN/DM is of bigger probability to contain same entities than in NYT50.
- 74% entities in CNN/DM can be linked to YAGO2, almost two times as much as in NYT50.
- Most importantly, sentence-entity edges in CNN/DM are much denser than in NYT50. *SE.Density* is defined as follows. Suppose the sentence-entity graph constructed from a document has *SE.Count* sentence-entity edges, *M* sentences, and *N* entities, then *SE.Density* of the document is calculated as $\frac{SE.Count+1}{M+N}$. Here other two types of edges are not counted. *SE.Density*>=1 means that the sentence-entity bipartite graph is a connected graph, and *SE.Density*<1 means that the sentence-entity bipartite graph is not a connected graph. Bigger *SE.Density*, more connected the bipartite graph.

### 4.2 Implementation and parameter settings

The model is implemented with Tensorflow in Python. The code will be released on Github under the link https://github.com/jingqiangchen/kbsumm. Due to the limited computational resource, the word vocabulary is limited to 40K words, and are initialized with 128-dimensional word2vec embeddings [57]. The entity vocabulary of CNN/DM is limited to 500K entities because there are more than 1000K entities in the dataset, and entities out of the vocabulary are replaced with the special UNK entity. The entity vocabulary of NYT50 is set as 146894 which is the whole entities recognized in NYT50. The entity vocabulary is initialized with 128-dimensional RDF2Vec embeddings [23]. One layer of the GRU cell is adopted as the RNN cell. The number of R-HGNN levels is set as 2. The dimension of graph node embeddings is set as 512. The dimension of the hidden state of the BiRNN encoder is set as 256. The dimension of the hidden state of the RNN decoder is 512. The parameters of Adam are set as those in [58]. The batch size is set to 15. Convergence is reached within 50K~80K training steps. It takes about one day for training 30k ~ 40k steps on a GTX-1080 TI GPU card.

For the multi-task selector, the top-k ranked sentences are selected as the extractive summary. k is set as 4 for CNN/DM and is set as 3 for NYT50, because the average sentence number in summaries of CNN/DM is 3.75 and of NYT50 is 2.94 as shown in Table 1. Similarly, the top-4 ranked entities for CNN/DM and the top-3 ranked entities for NYT50 are selected as salient entities according to the average number of entities in summaries of the two datasets as shown in Table 1. The input document is truncated into a maximum length of 100 sentences. The entity set of the input document is truncated into a maximum of 100 entities. For the entity-focused generator, the input text is truncated into a maximum length of 150 words. The decoding steps of the generator are limited to 100 steps.

**4.3 Comparing with existing methods**

More than ten baselines are compared with the proposed methods on CNN/DM and NYT50. The following are two versions of the proposed method and three strong existing baseline methods. Other baselines will be introduced when analyzing the evaluation results of the datasets.

- **RHGNNSumExt** The proposed extractive summarization method in this study by removing the entity-focused generator from the proposed model. Top-*k* ranked sentences are extracted as the extractive summary where k depends on datasets.
- **RHGNNSumAbs** The proposed complete model by first extracting top-k ranked sentences with RHGNNSumExt and then rewriting the sentences to an abstractive summary through the entity-focused generator, with a RL connector to connect the selector and the generator.
- **HGNNSum** This is the extractive summarization approach proposed in [2]. It constructs a bipartite sentence-word graph for an input document, where sentence-word edges are built if sentences contain words. HGNNSum applies the heterogeneous GNN to the graph and directly selects salient sentences as the summary by a sentence selector.
- **SENECA** This is the abstractive summarization approach proposed in [3] driven by entities to generate coherent summaries. Entities are first used to select salient sentences, and then a RL-based abstract generation module using coherence, conciseness and clarity as rewards is applied to compress the sentences to generate final summaries.
- **ASGARD** This is the abstractive summarization approach proposed in [27]. It constructs an entity-entity graph from the input document, where nodes are entities extracted from the document, and edges are predicates. The entity-entity graph is encoded with GNN, and the document is encoded with LSTM. And then the abstractive summary is generated by attending to graph encodings and document encodings. The constructed entity-entity graph is not linked to external knowledge graphs, and sentence-entity relations are not considered in the work. Two versions of ASGARD without reinforcement learning are compared. ASGARD-DOC treats the input document as a whole for encoding, while ASGARD-SEG segments the input document into a set of paragraphs which are encoded dependently and are then combined with LSTMs.

The following give the evaluations and analysis on CNN/DM and NYT50 respectively.

**Results on CNN/DM** Table 2 shows the evaluation results on the CNN/DM dataset. The first part is the LEAD-3 baseline which simply selects the leading three sentences from documents, and the ORACLE upper bound which is the ground truth extractive summary obtained as described in Section 3. The second part is four extractive summarization methods published recent years. The third part includes four state-of-the-art abstractive summarization methods including the pointer-generator network, the sentence rewriting method, the RL-based method, and the bottom-up method. The last part includes the two strong baselines SENECA and HGNNSum, and the proposed methods in this study.

**Table 2** Comparison results on CNN/DM using rouge F1 at the full summary length

| Method | Rouge-1 | Rouge-2 | Rouge-L |
|---|---|---|---|
| LEAD-3 [49] | 40.34 | 17.70 | 36.57 |
| ORACLE [59] | 52.59 | 31.24 | 48.87 |
| JECS [60] | 41.70 | 18.50 | 37.90 |
| LSTM+PN [42] | 41.85 | 18.93 | 38.13 |
| HER w/o Policy [60] | 41.70 | 18.30 | 37.10 |
| HER w Policy [61] | 42.30 | 18.90 | 37.60 |
| PG [49] | 39.53 | 17.28 | 36.38 |
| SENTREWRITE [45] | 40.88 | 17.80 | 38.54 |
| DEEPREINFORCE [62] | 41.16 | 15.75 | 39.08 |
| BOTTOM-UP [47] | 41.22 | 18.68 | 38.34 |
| SENECA [3] | 41.52 | 18.36 | 38.09 |
| ASGARD-DOC [27] | 40.38 | 18.40 | 37.51 |
| ASGARD-SEG [27] | 40.09 | 18.30 | 37.30 |
| HGNNSum | 42.31 | **19.51** | 38.74 |
| RHGNNSumExt | **42.39** | 19.45 | **38.85** |
| RHGNNSumAbs | 41.63 | 18.45 | 38.00 |

Bold values indicate that the best results

According to Table 2, RHGNNSumExt achieves the highest Rouge-1 and Rouge-L scores among all the extractive and abstractive methods. In particular, RHGNNSumExt outperforms HGNNSum according to Rouge-1 and Rouge-L scores. The main difference between the two models is three-fold: 1) RHGNNSumExt builds a sentence-entity graph with weighted multi-type edges, while HGNNSumm builds a sentence-word bipartite graph with one type of edges; 2) RHGNNSumExt utilizes both edge weights and edge types in the propagation process for calculations of node encodings; 3) RHGNNSumExt injects external knowledge from YAGO2 into the graph through linking entities to the knowledge base, and trains the entity embedding to fit the entity-entity edges. As shown in Table 1, each sentence in CNN/DM contains average 2.14 entity mentions, and the density of sentence-entity edges is 0.94, so there are enough sentence-entity edges in the sentence-entity graphs of the CNN/DM dataset for message propagations of R-HGNN. Nevertheless, the improvement of RHGNNSumExt over HGNNSum is not very significant. This is mainly because entity linking is a hard job and suffers from missing linking or wrong linking as mentioned in section 3.1. As shown in Table 1, only 16.81 out of 22.59 entities for CNN/DM can be linked to YAGO2, and there can be wrong linking among the 16.81 linked entities. The state-of-the-art entity linking method AIDA-light achieves about 80% precision in a standard small-scale short news dataset with high-quality manually annotated entity mentions [21]. As for CNN/DM, the news articles are much longer, and the quality of the mention annotations is much lower.

The proposed abstractive method RHGNNSumAbs achieves the highest Rouge-1 score among all the abstractive methods. In particular, RHGNNSumAbs achieves the higher Rouge-1 score and higher Rouge-2 score than SENECA does. SENECA is a two-step entity-driven method which first selects salient sentences and then generates coherent abstractive summaries using entities as queries. RHGNNSumAbs and SENECA use different content selection methods. RHGNNSumAbs utilizes the graph-based model which models sentence relations and incorporates knowledge graphs for content selection. And SENECA employs a single-layer unidirectional LSTM to recurrently extract salient sentences. RHGNNSumAbs also outperforms ASGARD-DOC and ASGARD-SEG. ASGARD does not employ an extractor before generative summary generation. The quality of the extracted sentences greatly determines the quality of the generative summaries.

**Results on NYT50** Table 3 shows the evaluation results on the NYT50 dataset. As with the limited-length ROUGE recall used in [55] and [2], the extracted sentences are truncated to the length of the human-written summaries and the recall scores are used instead of F1. The first two lines of Table 3 are baselines reported by [55] and the next two lines are the LEAD-3 baseline and the ORACLE upper bound for extractive summarization reported by [2]. The second part and the third part report the performance of other non-BERT-based studies and the proposed models in this study respectively. Because SENECA is not reported in NYT50, Table 3 does not show the scores of SENECA.

**Table 3** Comparison results on NYT50 using Rouge recall at the summary length of 100 words, where the result of the PG model is copied from [59] and '-' means the original paper did not report the result

| Method | Rouge-1 | Rouge-2 | Rouge-L |
|---|---|---|---|
| First sentence [55] | 28.60 | 17.30 | - |
| First k words [55] | 35.70 | 21.60 | - |
| LEAD-3 | 38.99 | 18.74 | 35.35 |
| ORACLE | 60.54 | 40.75 | 57.22 |
| COMPRESS [55] | 42.20 | 24.90 | - |
| SUMO [63] | 42.30 | 22.70 | 38.60 |
| PG [49] | 43.71 | 26.40 | - |
| DRM [62] | 42.94 | 26.02 | - |
| HGNNSum | 46.89 | 26.26 | 42.58 |
| RHGNNSumExt | 45.80 | 25.89 | 40.26 |
| RHGNNSumAbs | 44.30 | 24.33 | 37.03 |

The proposed model RHGNNSumExt outperforms most baselines on the NYT50 dataset. RHGNNSumExt does not outperform HGNNSum, because there are much less sentence-entity edges and much less entity mentions for NYT50 than for CNN/DM as shown in Table 1, making there are not enough sentence-entity edges for NYT50 for message propagations of R-HGNN. Concretely, sentences in NYT50 contains averagely 0.88 entity mentions and the *SE.Density* value is 0.57, respectively much less than 2.18 and 0.94 in CNN/DM. Moreover, the average number of the entities successfully linked to YAGO2 in documents of NYT50 is 8.75, also much smaller than 16.81 for CNN/DM, making there are less entity-entity edges for NYT50 because entity-entity edges are built if two entities co-occur on Wikipedia Web Pages. The detailed discussion of *SE.Density* will be given in the following subsection.

Nevertheless, the proposed abstractive method RHGNNSumAbs outperforms all abstractive baselines and most extractive baselines in terms of Rouge-1 scores in Table 3. In particular, RHGNNSumAbs achieves the higher Rouge-1 score than the pointer-generator network (PG) does. RHGNNSumAbs is a competitive abstractive summarization method for the NYT50 dataset.

**4.4 Discussion on density of sentence-entity edges**

As shown in the preceding subsection, the proposed method performs better on CNN/DM of higher *SE.Density* values, and performs worse on NYT50 of lower *SE.Density* values, indicating that *SE.Density* of datasets influences the performance of the proposed method. To see how the proposed method performs on datasets with different *SE.Density* values, and what values of *SE.Density* are suitable for usage of the proposed method, experiments are carried out on subsets of the two datasets with different *SE.Density* values.

Fig. 5 shows sample distributions on CNN/DM and NYT50 under different *SE.Density* values. As shown, sample distributions on the two datasets are rather different. By summing up the numbers of samples whose *SE.Density* are smaller than 0.7 in Fig .4, there are in total 79470 out of 103656 samples from NYT50. In contrast, there are in total 244485 out of 311491 samples from CNN/DM whose *SE.Density* is greater than 0.7. There are 25 samples in CNN/DM whose *SE.Density* is between 0.0 and 0.1, comparing to 520 in NYT50. And there are 123247 samples in CNN/DM whose *SE.Density* is bigger than 1.0, comparing to 2147 in NYT50. As a result of the distributions, the average *SE.Destity* of CNN/DM is 0.94 and the average *SE.Density* of NYT50 is 0.57.

For CNN/DM, four sub-datasets are created for comparison by selecting the samples of which *SE.Density* values are smaller than 0.5, 0.6, 07, and 0.8 respectively. For NYT50, four sub-datasets are created for comparison by selecting the samples of which *SE.Density* values are greater than 0.4, 0.5, 06, and 0.7 respectively. For each sub-dataset, the split of train, dev and test follows the original datasets.

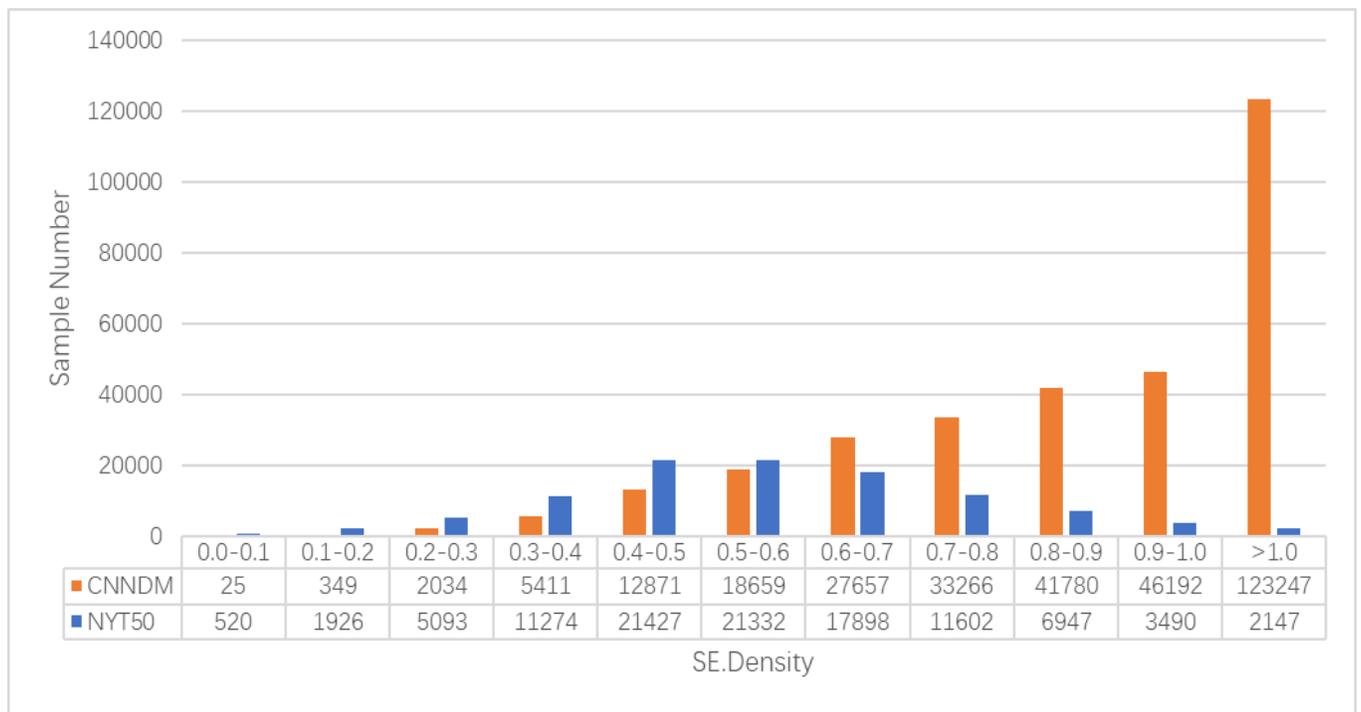

F**ig. 5** Sample distributions on CNN/DM and NYT50 under different *SE.Density* values

The proposed method RHGNNSumExt and the baseline method HGNNSum are trained and evaluated on the sub-datasets. For HGNNSum, the default settings provided in [2] is used, and the best eval models are adopted for test. The results are shown in Table 4 and Table 5.

**Table 4** Evaluation results on sub-datasets of CNN/DM with different *SE.Density* values

| SE.Density | Method | Rouge-1 | Rouge-2 | Rouge-L |
|---|---|---|---|---|
| <0.5 | RHGNNSumExt | 40.28 | 18.34 | 36.72 |
|  | HGNNSum | 40.36 | 18.59 | 36.84 |
| <0.6 | RHGNNSumExt | 40.82 | 18.95 | 37.42 |
|  | HGNNSum | 40.76 | 19.19 | 37.38 |
| <0.7 | RHGNNSumExt | 41.55 | 19.75 | 38.22 |
|  | HGNNSum | 41.44 | 19.61 | 38.12 |
| <0.8 | RHGNNSumExt | 41.72 | 19.78 | 38.37 |
|  | HGNNSum | 41.46 | 19.72 | 38.15 |

**Table 5** Evaluation results on sub-datasets of NYT50 with different *SE.Density* values

| SE.Density | Method | Rouge-1 | Rouge-2 | Rouge-L |
|---|---|---|---|---|
| >=0.4 | RHGNNSumExt | 45.22 | 25.19 | 41.37 |
|  | HGNNSum | 46.03 | 25.82 | 41.99 |
| >=0.5 | RHGNNSumExt | 45.62 | 25.57 | 41.81 |
|  | HGNNSum | 46.37 | 26.04 | 42.29 |
| >=0.6 | RHGNNSumExt | 46.72 | 26.33 | 42.81 |
|  | HGNNSum | 47.23 | 26.72 | 43.10 |
| >=0.7 | RHGNNSumExt | 47.56 | 27.16 | 43.71 |
|  | HGNNSum | 47.53 | 27.15 | 43.69 |

As shown in Table 4, for CNN/DM, the proposed method RHGNNSumExt performs worse than HGNNSum on the sub-datasets with *SE.Density*<0.5 and *SE.Dentity*<0.6, and RHGNNSumExt performs better than HGNNSum on the sub-datasets with *SE.Density*<0.7 and *SE.Dentity*<0.8. For CNN/DM, the performance of RHGNNSum improves when the *SE.Density* of sub-datasets increases.

As shown in Table 5, for NYT50, RHGNNSumExt performs better than HGNNSum on the sub-dataset with *SE.Density*>=0.7, and RHGNNSumExt performs worse than HGNNSum on the sub-datasets with *SE.Density*>=0.4, *SE.Dentity*>=0.5, and *SE.Dentity*>=0.6. Note that the average *SE.Density* of the entire NYT50 dataset is 0.57, and RHGNNSumExt performs worse than HGNNSum on the entire NYT50 dataset. For NYT50, the performance of RHGNNSum also improves when the *SE.Density* of sub-datasets increases.

### 4.5 Ablations

To better understand the effectiveness of different modules of the proposed model, ablation studies are conducted. Table 6 and Table 7 show the results on CNN/DM and NYT50 respectively.

**Effects of entity-level entity embeddings**. As introduced in Section 3.1, entity-level entity embeddings and word-level entity embeddings are concatenated as embeddings of entities. Entity-level entity embeddings partly represent external knowledge from the knowledge base YAGO2 in two ways: 1) entity-level entity embeddings are initialized on YAGO2 using RDF2Vec, and 2) the embeddings are trained with RHGNNSumExt by supervising with entity-entity edges constructed based on YAGO2.

To see the effect of adding entity-level entity embeddings, the embeddings are remove from the model, and the results are shown in the second lines of Table 6 and Table 7. The scores of RHGNNSumExt are higher than the model without entity embeddings, indicating adding entity-level entity embeddings improve the proposed summarization model.

**Table 6** Ablation studies on CNN/DM

| Method | Rouge-1 | Rouge-2 | Rouge-L |
|---|---|---|---|
| RHGNNSumExt | 42.39 | 19.45 | 38.85 |
| w./o. Entity-Level Entity Embeddings | 42.28 | 19.35 | 38.69 |
| w./o. Entity Embedding Supervising | 42.26 | 19.32 | 38.66 |
| w./o. Edge Weights | 42.33 | 19.39 | 38.76 |
| w./o. Edge Types | 42.25 | 19.24 | 38.62 |
| with Agg.Mean | 42.30 | 19.36 | 38.73 |
| w./o. EE Edges & SS Edges | 42.22 | 19.32 | 38.67 |
| RHGNNSumAbs | 41.63 | 18.45 | 38.00 |
| w./o. RL Connector | 41.55 | 18.36 | 37.91 |

**Table 7** Ablation studies on NYT50

| Method | Rouge-1 | Rouge-2 | Rouge-L |
|---|---|---|---|
| RHGNNSumExt | 45.80 | 25.89 | 40.26 |
| w./o. Entity-Level Entity Embeddings | 45.64 | 25.66 | 40.12 |
| w./o. Entity Embedding Supervising | 45.61 | 25.55 | 40.02 |
| w./o. Edge Weights | 45.69 | 25.79 | 40.22 |
| w./o. Edge Types | 45.62 | 25.65 | 40.11 |
| with Agg.Mean | 45.76 | 25.68 | 40.11 |
| w./o. EE Edges & SS Edges | 45.54 | 25.57 | 39.99 |
| RHGNNSumAbs | 44.30 | 24.33 | 37.03 |
| w./o. RL Connector | 44.24 | 24.28 | 36.94 |

To see the effect of supervising embeddings with entity-entity edges, this sub-objective is removed from the model, and the results are shown in the third lines of Table 6 and Table 7. The performance of the model without entity-entity edges supervision performs worse than the model with entity-entity edges supervision. Entity-entity edges supervision is to conserve entity-entity relatedness in entity embedding, and can improve the proposed model.

**Effects of edge weights and edge types in R-HGNN**. There are three types of edges with different weights in the sentence-entity graph. R-HGNN utilizes both edge weights and edge types in the propagation process for calculations of node encodings, combining the advantages of both the traditional GNN and the traditional R-GNN.

To see the effect of edge weights, the weights are removed from the graphs and the traditional R-GNN is applied instead of R-HGNN. As shown in the fourth lines of Table 6 and Table 7, using R-GNN without edge weights perform not as well as using R-HGNN with edge weights, indicating that edge weights on R-HGNN can improve the performance.

To see the effect of edge types, edge types are removed from the graphs and the traditional GNN is applied instead of R-HGNN. As shown in the fifth lines of Table 6 and Table 7, GNN not considering edge types performs not as well as R-HGNN considering edge types, indicating that edge types on R-HGNN can improve the performance.

**Effects of aggregating algorithms**. To see the effect of aggregating algorithms for the proposed model, the mean aggregating algorithm of HinSAGE [41] is used instead of the aggregating algorithm of R-HGNN by neglecting edge weights in the propagation process. As shown in the sixth lines of Table 6 and Table 7, RHGNNSumExt with Agg.Mean performs not as well as RHGNNSumExt, indicating that the aggregating algorithm of the original GNN which can make use of edge weights is more suitable than the mean aggregating algorithm.

**Effects of EE edges and SS edges**. One difference between the sentence-entity graph constructed in this study and the sentence-word graph constructed in [2] is that the entity-entity edges and the sentence-sentence edges are added to the sentence-entity graph besides the sentence-entity edges, while the sentence-word graph only has sentence-word edges. The seventh lines of Table 6 and Table 7 show the performance of the proposed model without EE edges and SS edges. The model without the two types of edges performs worse than the model with the two types of edges, and even worse than the model without entity embeddings. EE edges and SS edges are necessary for the sentence-entity graph.

**Effects of the RL connector**. The last two lines of Table 6 and Table 7 show the Rouge scores of the abstractive models with or without the RL connector respectively. RHGNNSumAbs outperforms RHGNNSumAbs without the RL connector. Note that RHGNNSumAbs without the RL connector directly applies the generator on the sentences selected by the multi-task selector RHGNNSumExt. The performance of the selector greatly affects the performance of the generator. The RL connector fine tunes the selector to better fit the generator, and improves the generative summaries.

# 5 Conclusions

This paper proposes an entity-guided text summarization framework by connecting Knowledge Graph and Graph Neural Network to make use of knowledge beyond text and cross-sentence relations in text for creating faithful summaries. The key components of the proposed summarization framework are the method of leveraging entities to connect GNN and KG, and the relational heterogeneous GNN for summarization. GNN is connected with KG by constructing the sentence-entity graph, and initializing and training entity embeddings based on KG. Concretely, external knowledge in KG is utilized in GNN through building entity-entity edges based on Wikipedia webpages, initializing entity node encodings in YAGO2, and supervising entity node encodings with entity-entity edges. The relational heterogeneous GNN calculates node encodings of the sentence-entity graph with weighted multi-type edges by combining the advantages of both the traditional GNN and the traditional R-GNN to make use of edge weights and edge types in the propagation process. Experiments carried out on CNN/DM show the proposed extractive summarization method outperforms all reported baselines without pre-trained language models. Experiments carried out on NYT50 show the proposed method outperforms most reported baselines. Experiments on sub-datasets of CNN/DM and NYT50 show that density of sentence-entity edges of constructed sentence-entity graphs greatly influence the performance of the proposed model. The greater the density, the better the performance of the proposed model. Ablation studies show effectiveness of the proposed method, and that R-HGNN outperforms the traditional GNN and the traditional R-GNN in making use of both edge weights and edge types for summarization. The results provide a promising step for other NLP tasks such as multi-modal summarization, question answering, and news image captioning to make use of cross-sentence relations in documents and external knowledge in KG.

**Acknowledgements** This research was sponsored by the National Natural Science Foundation of China (No.61806101).

**Availability of data and materials** The datasets and codes are available at https://github.com/jingqiangchen/kbsumm.